\documentclass[runningheads]{llncs}
\usepackage[T1]{fontenc}
\usepackage{booktabs}
\usepackage{graphicx}
\usepackage{subcaption}

\usepackage{draftwatermark}
\SetWatermarkText{Author Accepted Manuscript}
\SetWatermarkAngle{45}
\SetWatermarkLightness{0.95}

\begin{document}

\title{
Co-Creating Buildable and Open Social Robot Study Companions with University Students 
\thanks{This work was in part co-funded by the European Union and the Estonian Research Council grant (PRG3237).}
}
\titlerunning{Co-Creating Buildable Open Social Robots with University Students}

\author{
Farnaz Baksh\inst{1,2} \and
Matevž B. Zorec\inst{1,2} \and
Feiazie Baksh\inst{2} \and
Karl Kruusamäe\inst{1}}
\authorrunning{F. Baksh et al.}

\institute{
University of Tartu, Estonia \and
University of Guyana Robotics Club, Guyana\\
\email{farnaz.baksh@ut.ee}}

\maketitle
\begin{center}
\scriptsize
\textcolor{gray}{This is the author’s version of the work accepted for publication at 18th International Conference on Social Robotics (ICSR + ART 2026)}
\end{center}

\begin{abstract}
Open-source social robots offer accessibility, repairability, and student empowerment, yet the build itself often presents a barrier.
Existing platforms either ship pre-assembled, foreclosing hands-on learning, or expose students to unfamiliar fasteners, opaque wiring, and inaccessible service points that erode engagement.
Whether targeted mechanical redesign can lower this barrier whilst maintaining structural integrity remains untested.
Here we show that Design for Assembly (DfA) and Design for Disassembly (DfD) interventions reshape how a build feels before they shorten how long it takes.
Working with university students in Guyana and Estonia, we applied the Double Diamond framework to co-create the Robot Study Companion (RSC) v4.1: mapping pain points, then redesigning its chassis around twist-lock fasteners, snap-fit joints, and tool-free service latches.
Across two studies with developers and first-time builders, system usability climbed from \textit{Poor} to \textit{Excellent} (SUS 59.4 \textrightarrow\ 89.4), perceived workload trended downward (NASA-TLX 4.29 \textrightarrow\ 4.00), and mean assembly time trended downward (21.4 \textrightarrow\ 13.7 minutes, with juniors' learning effect), whilst orientation cues and navigation continuity for first-time builders emerged as the next documentation frontier.
Perceived workload, not completion time, appears to govern whether students take up open hardware.
\end{abstract}

\section{Introduction}
Social robots have shown promise in education as responsive tutors and peers, enhancing motivation and engagement through interactive experiences\cite{ruzic_teaching_2024}.  
However, to meaningfully benefit learners, these robots should not be mere black-box tools. Designing with students, not just for them, empowering them as co-creators and builders, helps them express identity and build agency\cite{Li_2023}.
Open-source hardware (OSHW) furthers this objective by enabling low-cost, customisable robots that foster technical agency \cite{Shi2024-lk} and constructionist learning.

Beyond initial interaction, a robot's physical lifecycle (its buildability and ``DIY-ability'' \cite{diy-socialrobot2017}) shapes the human--robot relationship itself. 
The wave of consumer social robot shutdowns (e.g., Jibo, Anki’s Cozmo, and Moxie) stranded owners with no recourse, illustrating that without serviceability, the relationship dissolves the moment the manufacturer withdraws\cite{kamino_revival_2026}. 
To foster enduring relationships and combat obsolescence, we can adopt a ``lifecycle perspective'' \cite{sustain_sr2024} that treats maintenance and repair as core components of the user experience.  
Bridging the gap between social robot longevity and student empowerment requires the integration of Design for Assembly (DfA) and Design for Disassembly (DfD) principles. 
DfA focuses on simplifying product structures to reduce assembly time and complexity by evaluating the necessity of each component and optimising the ease of its handling, insertion, and fastening\cite{Boothroyd1996,bogue_dfa}. 
DfD ensures product designs support end-user disassembly for maintenance, component reuse, or recycling at end-of-life \cite{Vanegas2018-ep,bogue_dfd}. 
These principles treat ease of repair, modification, and upgrade as fundamental design requirements, extending the lifecycle perspective \cite{sustain_sr2024} to the initial build phase.
Taken together, this shift can empower students to evolve their robot whilst enhancing self-efficacy through transparent, hands-on exploration of the hardware. \\

The Robot Study Companion (RSC)\footnote{\url{https://rsc.ee/}} project embraces this ethos with a user-centred approach by encouraging students to co-develop their desktop-sized social robots. 
In physical terms, the RSC comprises an affordable, compact, 3D printed desktop robot built around a Raspberry Pi with peripherals enabling multimodal interaction (touchscreen, servo-driven flippers, an LED ring, microphones, speakers)~\cite{baksh_open-source_2024}. 
Pilot workshops showed that students preferred approachable, multimodal robots and engaged readily in sketching ideas using a low-barrier one-slide design method~\cite{Baksh2024-dj}. 
Having evolved through several chassis iterations (v1\,\textrightarrow\,v4\,\textrightarrow\,v4.1), the platform supports university students facing isolation and motivational challenges by providing emotional support and a sense of physical presence~\cite{calafa_emotive_2025}. 

\par
In this paper, we present the complete co-creation cycle of the RSC v4.1, by moving from initial user engagement to empirical validation.
We ask: \textit{To what extent do DfA- and DfD-informed mechanical changes improve the perceived buildability, serviceability, and workload of an open-source social robot, as experienced by both developers and first-time builders?}

To address this, we apply the Double Diamond framework~\cite{designcouncilHistoryDouble} to the RSC v4.1 redesign, contributing:
\begin{enumerate}
    \item an end-to-end co-design methodology mapping user-identified pain points to targeted hardware changes (Section \ref{sec:methods})
    \item open-source hardware artefacts for the RSC v4.1 chassis
    \item empirical evidence that DfA/DfD principles reduce perceived workload and raise usability for both developers and first-time builders (Section \ref{sec:results-testing})
    \item a reframing of buildability and serviceability as social affordances of the human--robot relationship (Section \ref{sec:discussion})
\end{enumerate}

\section{From User Insights to a More Buildable Robot}
\label{sec:methods}
\begin{figure} [htbp]
\centering
\includegraphics[width=0.82\linewidth]{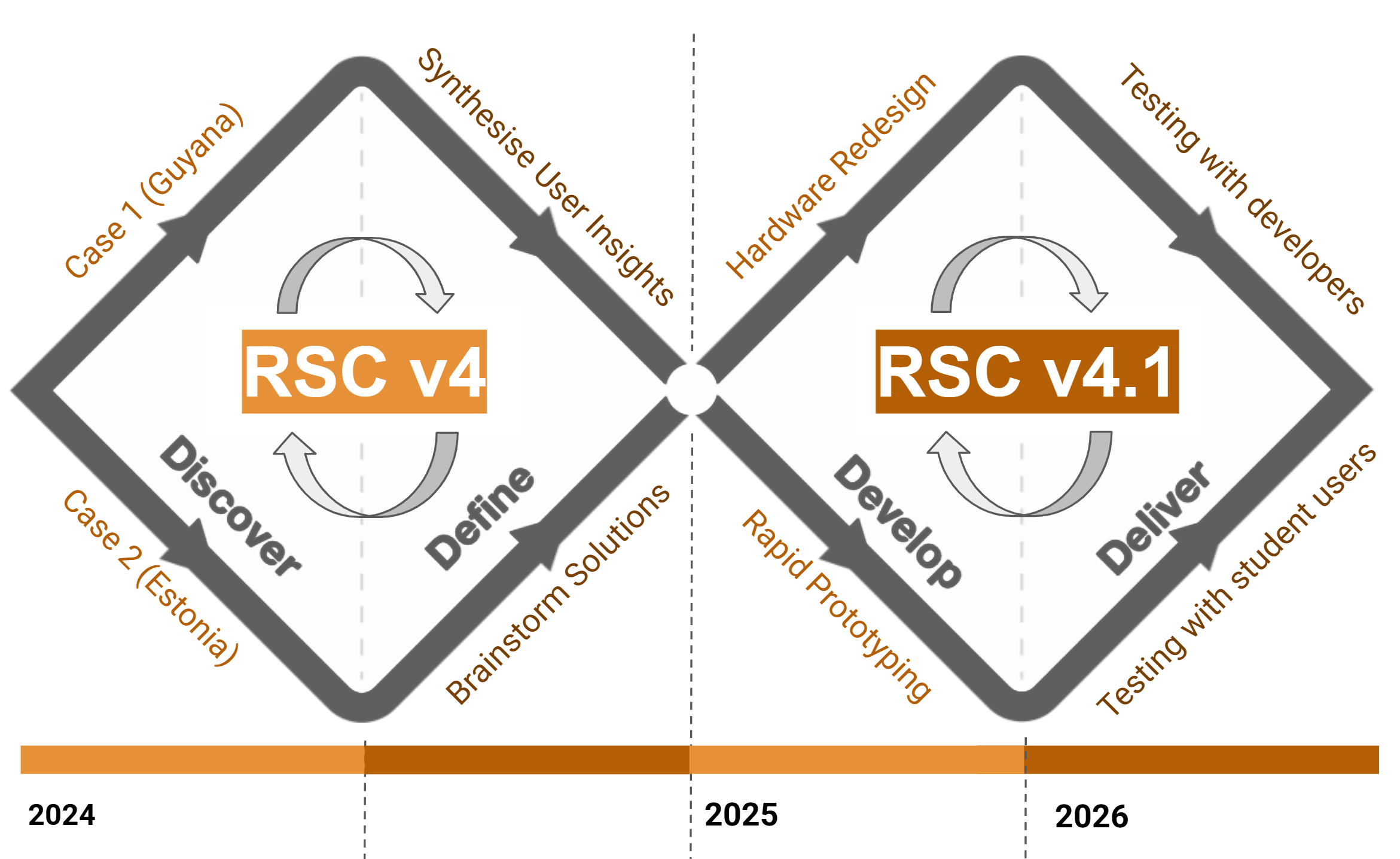}
\caption{Co-creating the RSC using the Double Diamond approach with university students (2024--2026). 
The first diamond (2024) derives design requirements from two replication cases in Guyana and Estonia (Discover/Define, §\ref{sec:first-diamond}). 
The second (2025--2026) implements targeted hardware redesigns guided by DfA/DfD principles (Develop, §\ref{sec:develop}) and evaluates them through a within-team benchmark and a first-time-builder user study (Deliver, §\ref{sec:deliver}).}
\label{fig:rsc-dd}
\end{figure}

We follow a mixed-methods, iterative co-design approach structured by the Double Diamond~\cite{designcouncilHistoryDouble} (Fig.~\ref{fig:rsc-dd}). 
This framework separates problem-framing (Discover, Define) from solution-shaping (Develop, Deliver) across two iterative cycles. 
We pair this with established Design for Assembly~\cite{Boothroyd1996,bogue_dfa} and Design for Disassembly~\cite{bogue_dfd,dfd_2015,dfa-d_2017} guidelines to translate student feedback into tangible hardware interventions.

\subsection{First Diamond: Discover \& Define}
\label{sec:first-diamond}
To define the hardware requirements for RSC v4.1, we summarise two replication studies using RSC v4 chassis design \cite{baksh_open-source_2024}. 
These cases reveal the real-world friction between open-source documentation and the physical assembly process. 

\textbf{Case 1 (Guyana):}
Two students at the University of Guyana (UG) replicated the RSC v4 as part of a capstone computer science project \cite{rie_2025}. 
With high import duties and limited access to tools (e.g.\ soldering irons, 3D printers), the students relied on local support from the UG Robotics Club. 
Despite these hurdles, the students completed the build over several days and expressed significant pride in building their first robot from scratch, proposing feature improvements such as LED notifications.

\textbf{Case 2 (Estonia):}
We hosted a one-day workshop with four computer engineering students at the University of Tartu, using a fully equipped laboratory.
The goals were to provide a hands-on introduction to the RSC and gather feedback for developer community events.
Despite the availability of all tools, the build proved time-consuming for same-day completion.
This collaborative build helped us prioritise design improvements, highlighting the importance of designing the RSC for ease of buildability and maintenance, whilst peer learning (experienced participants teaching others to solder) demonstrated the platform's potential as a social robotics learning tool.

\begin{table}[!h]
\caption{Comparison of RSC replication case studies (v4 chassis) -- Synthesising user insights and resulting design implications for v4.1}
\label{tab:cases}
\centering
\footnotesize
\setlength{\tabcolsep}{6pt}
\begin{tabular}{p{2.1cm}p{2.6cm}p{2.6cm}p{3.2cm}}
\toprule
\textbf{Category} & \textbf{Guyana Case} & \textbf{Estonia Case} & \textbf{Design Implication} \\
\midrule
Tooling & Limited access & Readily available & Minimise required tools \\
\midrule
Assembly Challenges & Component \newline integration and soldering precision & Wire routing and screw access & Modular wiring and improved mechanical access \\
\midrule
Documentation Issues & Unclear instructions in assembly steps & --- & Improve assembly documentation \\
\midrule
Hardware Maintenance & SD card \newline inaccessible & \raggedright Sliding panels utilised & Prioritise serviceable layout \\
\midrule
Troubleshooting & \raggedright Required expert assistance & Fault diagnosis \newline  required partial disassembly & Design for easier debugging and modular diagnostics \\ 
\bottomrule
\end{tabular}
\end{table}

\paragraph{Synthesis of User Insights (Define Phase)}
We synthesised the observations from both case studies into the key design challenges summarised in Table \ref{tab:cases}, marking the transition from the Discover phase (observing real-world assembly experiences) to the Define phase (identifying concrete design requirements).
Based on the synthesis in Table \ref{tab:cases}, we identified three core problem areas: (1) \textbf{Complex assembly}, involving intricate wire management and hard-to-reach screws; (2) \textbf{Poor serviceability}, specifically the inaccessible SD card and the need for significant disassembly to diagnose simple faults; and (3) \textbf{Static documentation}, where ambiguous instructions increased assembly time. 
These findings broadened the design brief from assembly alone to the full lifecycle of build, repair, and maintain, forming the foundation for the subsequent hardware iterations.

\subsection{Second Diamond: Develop}
\label{sec:develop}
\begin{figure}
\centering
\includegraphics[width=\linewidth]{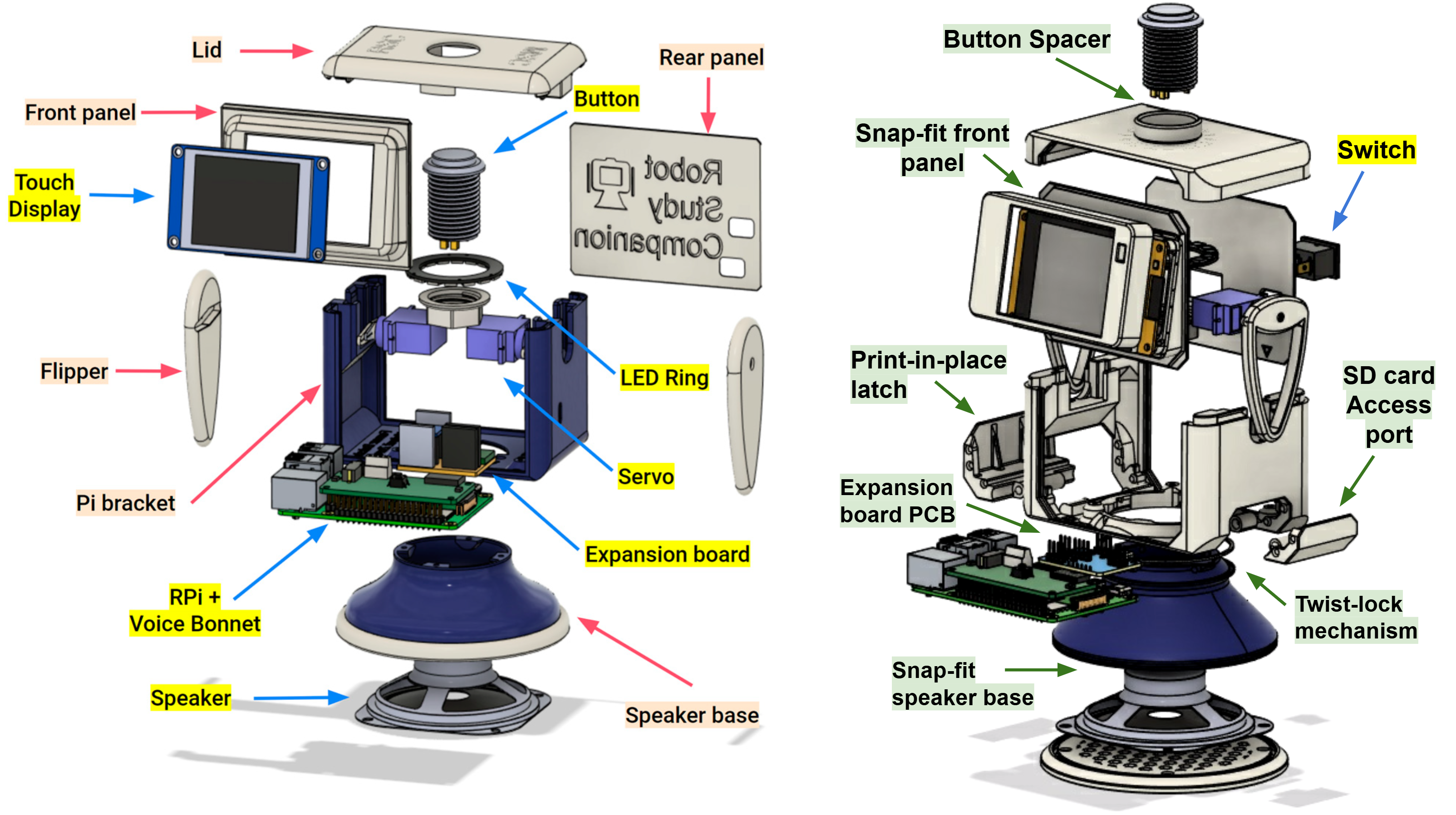}
\caption{Side-by-side comparison of the RSC version 4 (left) \cite{baksh_open-source_2024} and the updated version 4.1 prototype (right). 
The RSC project follows a modular versioning scheme, with independent version tracks for the chassis, electronics, and software. 
The designations v4 and v4.1 we use here refer specifically to the chassis mechanical design. 
We describe the overall platform release history on our project's GitHub: \url{https://github.com/RobotStudyCompanion/}}
\label{fig:rsc-compar}
\end{figure}

The Develop phase addressed these pain points through a hardware redesign (v4.1) guided by DfA and DfD principles, realised in the v4 \textrightarrow{} v4.1 chassis transition (Fig.~\ref{fig:rsc-compar}; Table~\ref{tab:dfa_dfd}).

\begin{table}[htbp]
\caption{DfA/DfD guideline \(\rightarrow\) v4.1 intervention crosswalk. Generic guidelines synthesised from~\cite{Boothroyd1996,bogue_dfa,bogue_dfd,dfa-d_2017,dfd_2015}; v4.1 changes shown alongside the user-identified problem each addresses (Table~\ref{tab:cases}).}
\label{tab:dfa_dfd}
\centering\footnotesize
\setlength{\tabcolsep}{2pt}
\begin{tabular}{p{1.0cm}p{4.0cm}p{6.4cm}}
\toprule
\textbf{Lens} & \textbf{Guideline} & \textbf{RSC v4.1 Hardware intervention} \\
\midrule
DfA & Reduce part count & Snap-fit speaker base and front panel replace multi-screw assemblies \\
DfA & \raggedright Built-in / standardised fasteners & Twist-lock chassis replaces internal screws \\
DfA & Mistake-proofing & One-way component fits and fiducial markers on printed parts \\
\addlinespace
\hline
\addlinespace
DfD & Modular architecture & Subassemblies removable independently \\
DfD & Standardised interfaces & Bespoke expansion PCB consolidates wiring breakouts \\
DfD & \raggedright Accessible disassembly points & Print-in-place latches give tool-free SD/USB access \\
DfD & Avoid encapsulated \newline components & Access ports allow Pi/peripheral inspection and usage \\
\addlinespace
\hline
\addlinespace
Docs & Step-by-step visual instructions; mistake-proofing cues & Rewritten guide at \url{https://rsc.ee/docs} with synchronised video and text \\
\bottomrule
\end{tabular}
\end{table}

Reinforced printed parts (reducing \textit{print-tolerance} variability) complement the three change categories: DfA (assembly friction), DfD (serviceability), and documentation rewrite (Table~\ref{tab:dfa_dfd}); Section~\ref{sec:results-testing} evaluates their effects.

\subsection{Second Diamond: Deliver -- Study Design}
\label{sec:deliver}

We evaluated the v4.1 redesign through two complementary studies, each targeting the three pillars identified in the Define phase (§\ref{sec:first-diamond}): buildability, serviceability, and perceived workload.

\textbf{Study~1} (Section~\ref{sec:study1}) was a within-subjects benchmark with four members of the RSC development team, comparing v4 and v4.1 across full assembly and disassembly cycles to isolate the effect of the mechanical redesign.
\textbf{Study~2} (Section~\ref{sec:study2}) extended validation to seven first-time student builders working with v4.1 only, testing whether the buildability gains generalised beyond the development team.

\textbf{Data Collection.} 
Both studies used the same instrument set to ensure descriptive comparability: timed assembly and disassembly tasks; NASA-TLX (1--10) for perceived workload; Single Ease Question (SEQ; 1--7) for task-specific difficulty; System Usability Score (SUS; 0--100) for usability; a nine-item documentation clarity Likert (1--5) covering visual clarity, step sequencing, parts identification, orientation cues, and navigation continuity; and structured qualitative debriefs with observer notes coded against themes.
We kept v4.1 documentation in an alpha state to test whether the redesigned hardware affordances could carry the build with thinner instructions.

\section{Results: Validation of RSC v4 \& v4.1}
\label{sec:results-testing}
We report descriptive statistics to characterise the direction and magnitude of changes across both studies (Table~\ref{tab:metrics}), using the data instruments described in §\ref{sec:deliver}: NASA-TLX~\cite{HART1988139}, SUS~\cite{brooke1996sus}, SEQ, and a nine-item documentation clarity Likert.
Study 1 benchmarked the RSC development team ($n$\,=\,4) across both versions; Study 2 assessed v4.1 with end users who had no prior exposure to the RSC platform \cite{baksh_open-source_2024}. 

\begin{table}[t]
\caption{Summary of key metrics across Studies 1 and 2.
Values report means with standard deviation in brackets: M\,(SD) where available.
Subscale and documentation rows report cell means only (no SD).
The v4.1 developer assembly mean incorporates each junior's second attempt; the first-attempt-only mean is 20.4 minutes (§\ref{sec:study1}).
\textsuperscript{\dag}n\,=\,3--4 (single partial response).
}
\label{tab:metrics}
\centering\footnotesize
\setlength{\tabcolsep}{3.5pt}
\begin{tabular}{lcccc}
\toprule
& \multicolumn{2}{c}{Study 1 (Devs: $n$\,=\,3--4\textsuperscript{\dag})}
& \multicolumn{1}{c}{Study 2 (Users: $n$\,=\,7)} \\
\cmidrule(lr){2-3} \cmidrule(lr){4-4}
\textbf{Metric} & \textbf{v4} & \textbf{v4.1} & \textbf{v4.1} \\
\midrule
Assembly (asm.) time (minutes)         & 21.4 (13.0) & 13.7 (5.3)  & 39.1 (7.5) \\
Disassembly (dis.) time (minutes)      & 5.0 (2.1)   & 4.7 (2.0)   & 7.5 (2.7)  \\
Assembly\,:\,disassembly ratio   & 4.3$\times$ & 2.9$\times$  & 5.2$\times$ \\
\addlinespace
NASA-TLX overall (asm.)     & 4.29 (2.47) & 4.00 (2.04) & 3.60 (1.28) \\
NASA-TLX overall (dis.)     & 3.21 (1.24) & 2.04 (0.92) & 2.83 (1.42) \\
\quad Mental Demand (asm.)  & 4.25        & 3.25         & 4.43        \\
\quad Physical Demand (asm.)& 4.00        & 3.00         & 3.57        \\
\quad Effort (asm.)         & 4.50        & 5.75         & 3.71        \\
\quad Frustration (asm.)    & 4.25        & 3.75         & 2.57        \\
\addlinespace
SEQ assembly (1--7)         & 4.50 (1.73) & 5.00 (1.41) & 6.17 (0.41) \\
SEQ disassembly (1--7)      & 5.75 (0.96) & 6.75 (0.50)  & 6.57 (0.53) \\
SUS (0--100)                & 59.4 (21.1) & 89.4 (16.4) & 74.6 (12.4) \\
\addlinespace
Doc.\ clarity overall (1--5)& 3.97       & 3.58         & 4.10        \\
\quad Orientation clarity   & 3.33\textsuperscript{\dag}        & 2.75         & 3.43        \\
\quad Navigation continuity & 4.25        & 4.25         & 3.43        \\
\bottomrule
\end{tabular}
\end{table}

\subsection{Study 1 -- Benchmarking with RSC Developers}
\label{sec:study1}
 
Four developers (two senior, two junior) performed a within-subjects comparison of v4 and v4.1.
Each participant performed a full assembly and disassembly of both versions in a single session lasting one to two hours. 
To capture learning effects, the two junior developers conducted a second v4.1 assembly attempt following a day or two after.
Three participants followed the task order v4.1 assembly, v4.1 disassembly, v4 assembly, v4 disassembly; one participant (P\textsubscript{2}) performed the tasks in the reverse version order (v4 first, then v4.1), providing informal counterbalancing.

\subsubsection{Task Completion and Timing.}
\begin{figure}[!h]
\centering
\includegraphics[width=\linewidth]{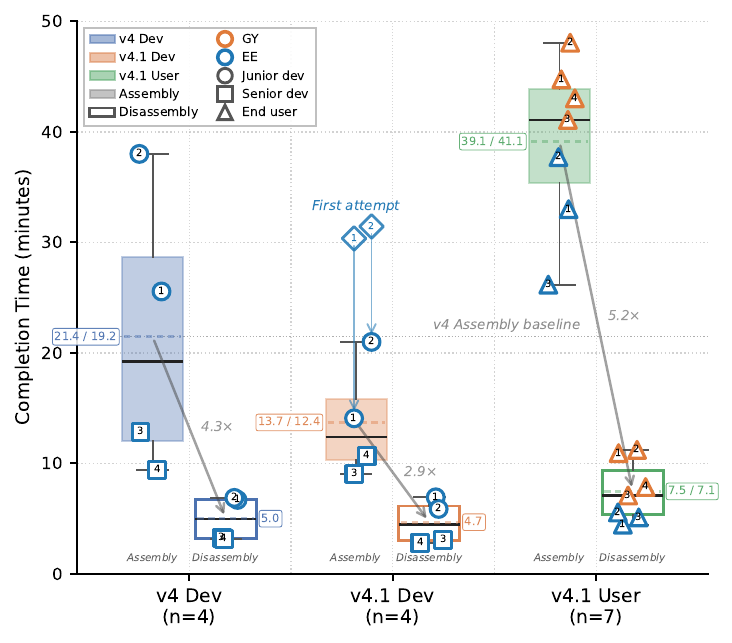}
\caption{Task completion times for both studies. 
Boxes indicate median, interquartile range, whisker (min/max) and mean (dashed tick) for paired assembly (filled) and disassembly (outlined) trials. 
Grey arrows show mean slope ratios (assembly \textrightarrow\ disassembly). 
Markers identify participant role and locale; open diamonds with arrows denote the learning effect from first to second v4.1 attempts for junior devs.}
\label{fig:fig_timing}
\end{figure}

Disassembly times were nearly identical across versions (4.7--5.0 minutes; Table~\ref{tab:metrics}), confirming that v4.1 preserved the DfD objective of rapid teardown. 
Assembly times trended downward (21.4 \textrightarrow\ 13.7 minutes; see Fig.~\ref{fig:fig_timing}), compressing the assembly-to-disassembly ratio from 4.3× to 2.9×. 
Senior developers completed both assemblies roughly three times faster than juniors on first attempt (M\,=\,9.9 vs 30.9 minutes for v4.1). On a first-attempt-only basis, v4.1's mean is 20.4 minutes against v4's 21.4; the further reduction to the 13.7 minutes in Table~\ref{tab:metrics} reflects a substantial junior learning effect ($\approx$43\%, with one junior reaching 14.1 minutes on second attempt, Fig.~\ref{fig:fig_timing}), suggesting the redesign primes learning rather than driving the speedup on first build.

\subsubsection{Perceived Workload and Usability.}
The v4.1 redesign resulted in a rise in perceived usability with SUS scores (Fig.~\ref{fig:radars-Sus_docs}a) jumping from 59.4 (v4, \textit{Poor}) to 89.4 (v4.1, \textit{Excellent}).
Improved tactile confidence rooted this shift; developers described v4.1 as ``robust'' (P\textsubscript{4}) and ``solid'' (P\textsubscript{2}), noting that the ``twist and lock [mechanism] feels confident [and] good while assembling''(P\textsubscript{3}).

\begin{figure}[!h]
\centering
\includegraphics[width=\linewidth]{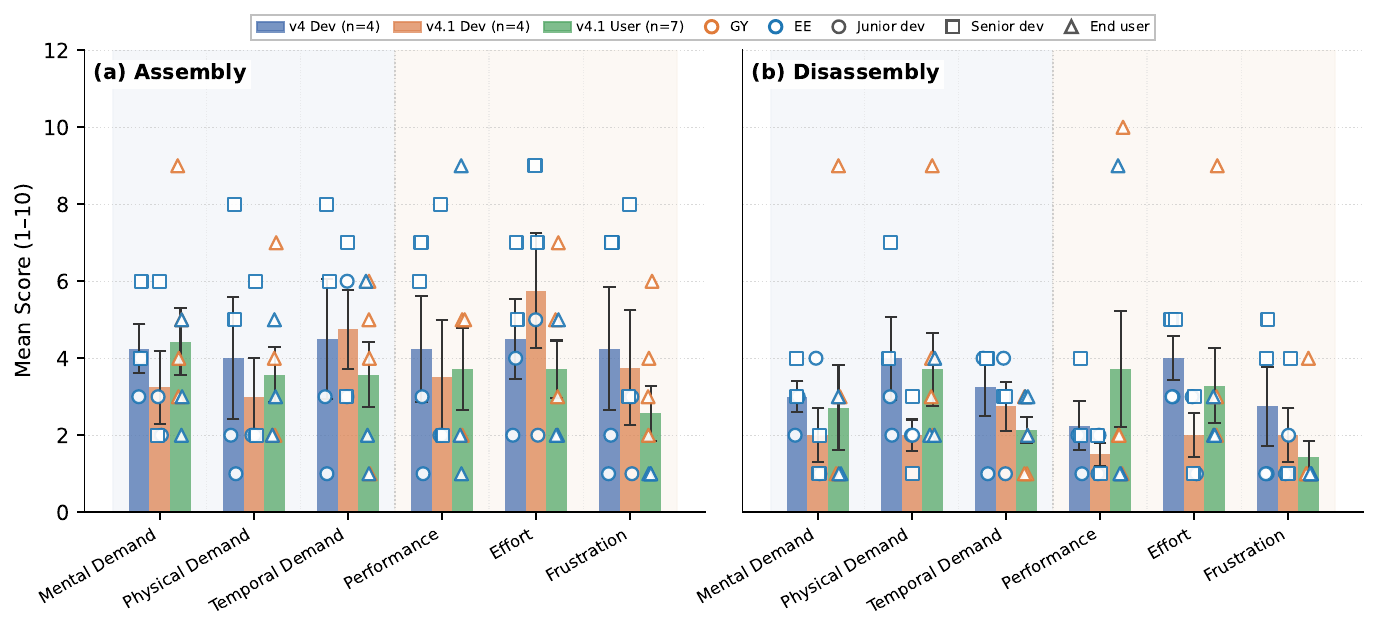}
\caption{NASA-TLX subscale scores for assembly (a) and disassembly (b), comparing v4 developers and v4.1 developers (Study~1, within-subjects) against v4.1 end users (Study~2). 
Bars give group means with $\pm 1$ standard error; dots show individual scores by site (EE/GY) and role. 
Lower scores are better throughout: less workload for Mental, Physical, Temporal, Effort and Frustration; better self-rated success for Performance (reverse-anchored, 1\,=\,perfect).}
\label{fig:fig4_tlx}
\end{figure}

Whilst overall NASA-TLX assembly scores remained stable, the subscales reveal a shift in the workload profile (Table~\ref{tab:metrics}). 
Frustration dropped from v4 to v4.1 (M\,=\,4.25\,\textrightarrow\,3.75) as the redesign addressed the ``finicky'' (P\textsubscript{3}) and ``incomplete'' (P\textsubscript{2}) nature of the previous version.
The apparent rise in Effort (Fig.~\ref{fig:fig4_tlx}a) is driven by self-critical ratings from the two senior developer reviewing the design; thus we do not interpret the difference as a workload signal.
Primary pain points in v4 included tool-switching and inaccessible fasteners (``Screws! Took a lot of screwing'', P\textsubscript{4}; ``One place for a screw was hard to reach'', P\textsubscript{2}. 
P\textsubscript{1} noted that using ``only one kind of screws'' made v4.1 simpler to use and more accessible. 

For disassembly, workload reduction was consistent across all measures (Table~\ref{tab:metrics}, Fig.~\ref{fig:fig_timing}). SEQ scores reinforced this pattern, with v4.1 rated equal or easier than v4 for both tasks. P\textsubscript{4} shared that v4.1 ``feels easy to come apart'' and allowed for parallelising the teardown by instinctively knowing ``what to do next'' whilst P\textsubscript{1} said ``Everything comes easily apart.'' 

\subsubsection{Serviceability and Documentation.}
RSC v4.1 achieved its goal of tool-free serviceability. 
Participants confirmed that the print-in-place latches allowed immediate access to the SD card and USB ports, tasks deemed ``impossible'' on v4 without full disassembly.

\begin{figure}[!h]
\centering
\includegraphics[width=\linewidth]{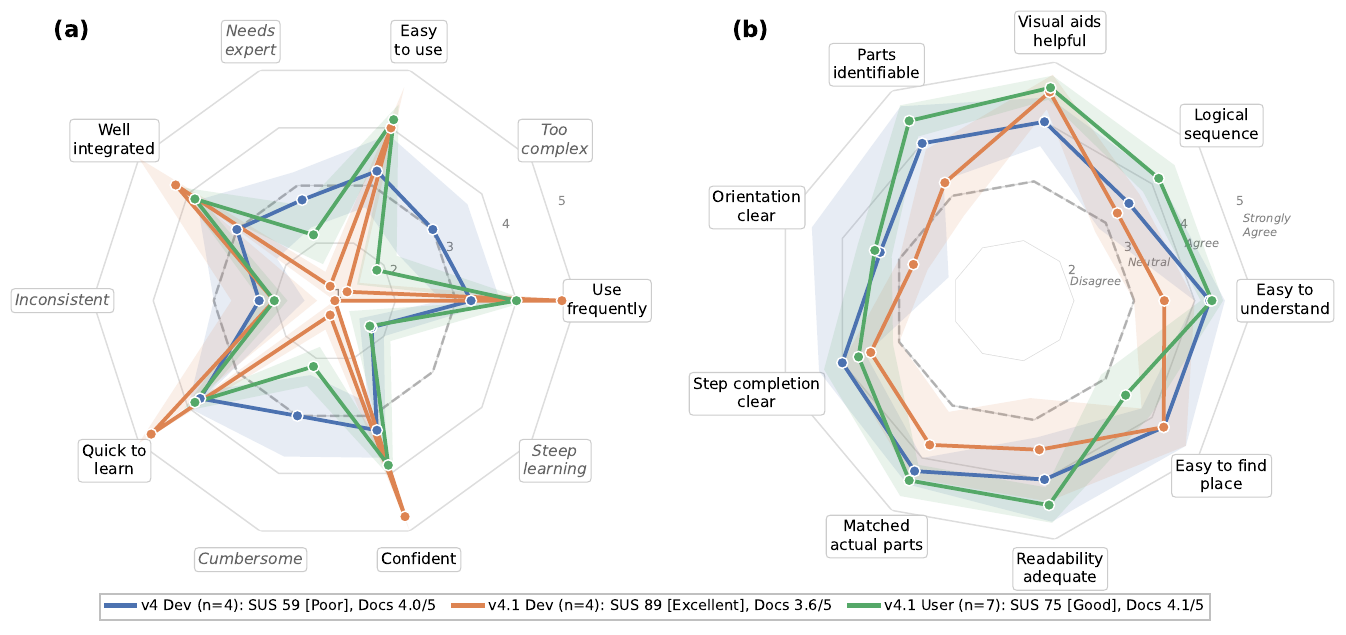}
\caption{Item-level perceptual breakdowns by condition. \textbf{(a) SUS per-item means}, upright labels = positive items, italicised = reverse-coded (lower\,=\,better).
The star shape reflects consistent polarity engagement. \textbf{(b) Documentation clarity (1--5)}, administration order; inner zone below~2 masked. 
Dashed contour at~3 marks Neutral; bands show $\pm 1$ standard error around each group mean (narrower bands = tighter estimate).
Legend gives SUS and doc-clarity means.}
\label{fig:radars-Sus_docs}
\end{figure}

However, v4.1 scored lower on documentation clarity than v4 ($M=3.58$ vs $M=3.97$; Fig.~\ref{fig:radars-Sus_docs}b). As the v4.1 guide was an intentionally incomplete alpha, we do not interpret the cross-version delta as evidence about v4.1 documentation quality; we treat it as a manipulation check confirming the instrument detects clarity differences. 
Specifically, orientation (wiring) clarity remained the weakest item ($M=2.75$), with P\textsubscript{2} noting: ``[doc] video was good enough [but] I didn't understand which way the wires go.'' 

The documentation format created a ``pause-act-rewind'' friction among the senior developers with P\textsubscript{4} noting ``watching videos slow[ed] me down, knowing what [I was] doing and feel [I] could finish assembly faster [... felt] like it took more effort''. 
Whilst P\textsubscript{3} said ``[it] felt faster to just put the 
robot together rather than go back and forth with docs.'' P\textsubscript{3} also described the process as ``frictiony'', suggesting that the UI needs better controls, such as using ``arrow keys to quickly go back and forth'' or the ``spacebar to pause/play''.

\subsection{Study 2 -- End User Validation}
\label{sec:study2}
Seven students from the University of Guyana ($n=4$) and the University of Tartu ($n=3$) completed the validation of the RSC v4.1. 
Participants represented diverse technical backgrounds (Computer Science, AI, and Engineering), and whilst most had used hand tools, none had prior exposure to the RSC platform. 
This study focused on whether the DfA/DfD gains observed in Study 1 generalised to first-time builders.
Participants generally found the process ``quite fun'' (P\textsubscript{4}-GY), noting it was ``a valuable way to learn about the [RSC] parts''.
P\textsubscript{2}-EE described the build as ``visual[ly] nice, 
playful and looks interesting. Overall it is fun and engaging 
to assemble (much like legos)'', praising the ``twist and lock -- 
mechanical parts really good and well design[ed], fits quite snug.''

\subsubsection{Task Completion and Perceived Workload.}
Mean assembly time was 39.1 minutes (SD\,=\,7.5) and disassembly 7.5 minutes (SD\,=\,2.7) (see Table \ref{tab:metrics}); first-time builders' heavier documentation consultation likely drives the 5.2× ratio against developers' 2.9--4.3× (Fig.~\ref{fig:fig_timing}).
Despite this, NASA-TLX scores reflected a low-to-moderate workload with a marked drop in Mental Demand and Frustration (Fig.~\ref{fig:fig4_tlx}) during disassembly.
As P\textsubscript{2}-GY observed ``Disassembling is much quicker. I think anyone can disassemble without having instructions''. SEQ ratings corroborated this ease of use, with disassembly reaching a near-perfect ``Very Easy'' mean.
However, user Mental Demand during assembly remained higher than that of developers (Fig.~\ref{fig:fig4_tlx}), which participants attributed to the effort of identifying parts and reinforcing wiring identification. P\textsubscript{3}-GY suggested that ``[labelling] actual parts ... would be nice since it would reduce my mental load''. 

\subsubsection{Hardware Affordances and Serviceability.}  
All seven participants successfully accessed the RPi's SD/USB ports, though with minor discoverability gaps: one participant required a prompt to locate the SD card slot, and another initially missed the side-panel port latches. 
The tool-free redesign was a highlight; P\textsubscript{3}-EE noted the robot ``mostly relies on snap-fit so that way it was very easy to (dis)assemble'', whilst P\textsubscript{1}-EE praised the new print-in-place latches as ``much easier''.
Several participants reported concerns around snap-fit components, uncertain whether resistance signalled an error or correct insertion: ``Scared I would break things with the components that are meant to snap in'' (P\textsubscript{3}-GY).
This tactile fragility (absent among developers in Study~1) represents a feedback issue distinct from documentation clarity and specific to first-time builders.
It indicates that participants treated the partially-built robot as something worth protecting before any social interaction had taken place, a point we return to in Section~\ref{sec:discussion}. 
Despite minor discoverability issues and a dislike for ``touching the screen panel during removal'' (P\textsubscript{1}-EE), the overall SUS score of 74.6 (SD\,=\,12.4; Table \ref{tab:metrics}) places the hardware in the \textit{Good} range (Fig.~\ref{fig:radars-Sus_docs}a).

\subsubsection{Documentation Bottleneck and Wiring} 
Whilst six of seven participants identified video guidance as the most helpful feature, they concurrently cited wiring as the primary difficulty. 
Issues ranged from pin identification, with P\textsubscript{1}-EE and P\textsubscript{2}-GY noting the most confusing part was ``the wires, [and] what wire to connect to which pin'' and a heavy reliance ``on the colour coding'' (P\textsubscript{1}-EE)—to the physical challenge of ``fitting [my] fingers in the tiny space'' (P\textsubscript{2}-EE).  

The metrics reflect this cognitive load: whilst overall clarity remained high (Table~\ref{tab:metrics}), Orientation and Navigation (Easy to find place) scored lowest (Fig.~\ref{fig:radars-Sus_docs}b). 
These subscales capture ``workflow interruption'' scenarios where a user pauses the video to act and must then re-establish their position, which points to a structural flaw in the video-only format: ``a frustrating pause-act-rewind'' cycle. 
P\textsubscript{1}-GY observed that ``[s]tep-by-step diagrams would be nice to have alongside the videos'' to provide static anchors and avoid constant backtracking.
P\textsubscript{2}-EE further remarked on a tension between the robot's praised compact form and the resulting wiring friction, noting that ``whil[st] the design is good for user experience,'' ``assembly gets tricky (especially in the end)'' as wires must fit precisely inside the housing. 
To mitigate this, two participants suggested ``implementing hooks or compartments [...] so that the wires fit better''. 
Finally, participants identified two concrete documentation errors: ``a step-ordering flaw requiring unnecessary backtracking'' to remove the screen panel during wiring (P\textsubscript{1}-EE) and a screw-count discrepancy between video and written instructions (P\textsubscript{3}-GY).

\section{Discussion}
\label{sec:discussion}
The RSC project demonstrates that lowering buildability barriers can bridge socioeconomic divides, enabling students in diverse contexts (e.g., Guyana and Estonia) to engage with social robotics. 
By prioritising Design for (Dis)Assembly and open-source principles, whilst using affordable, off-the-shelf electronics and 3D-printable materials~\cite{baksh_open-source_2024}, we aim for a sustainable lifecycle that reduces e-waste and extends operational lifespan.
Findings from both studies converge on a central result: the v4.1 redesign made assembly feel more intuitive and usable.

\subsection{Reshaping the ``Feel'' of Assembly}
The integration of DfA and DfD principles shifted the robot's workload profile rather than just its assembly speed. 
Mechanical changes (twist-locks, one-way snap-fits) measurably reduced screw-related Frustration across both developer and end-user populations, whilst qualitative feedback consistently described the redesigned build as a shift from physical friction to mental modelling and sequencing.
This suggests that ``ease of use'' in robotics is not merely a reduction in labour but a shift in the kind of attention the build demands. 
The tactile experience was central to this shift; developers noted that the v4.1 twist-lock mechanism felt ``confident'' and ``good while assembling'' (P\textsubscript{3}). 
This was echoed by users who liked ``how the parts feel in my hands'' (P\textsubscript{4}-GY) and found the components a ``perfect size in terms of handling'' (P\textsubscript{3}-EE).
However, some reported fearing they would ``break things'' (P\textsubscript{3}-GY) when encountering the resistance of a snap-fit. 
One suggestion is that hardware feedback can be explicit: designers can provide \textit{success cues}, such as an audible click, to help users distinguish a secure fit from a potential break.

\subsection{Buildability as a Relational Precondition}
We argue that hands-on assembly is an interaction-shaping variable, not just a hardware prerequisite. 
Assembly establishes the user's stance toward the robot as a co-created artefact rather than a delivered product. 
Participants treated the robot as something worth protecting before any social interaction took place.
The ``like Lego'' engagement (P\textsubscript{2}-EE) and affective vocabulary (e.g. ``fun,'' ``engaging,'' ``cool'') suggest the build phase generates a positive stance that may persist into later use.
By removing the mechanical frustration of v4, which (P\textsubscript{2}) described as needing a ``3rd arm to hold stuff in place'', we allow the user to focus on the robot's agency. 
For educational OSHW platforms, the design goal should be optimising the first-attempt experience to foster psychological ownership and self-efficacy.
 
\subsection{Beyond the ``Pause-Act-Rewind'' Cycle}
Our findings confirm that documentation format is as vital as its content. 
Whilst six of seven Study~2 participants mentioned video guidance as most helpful, it also created a structural bottleneck of a repetitive ``pause-act-rewind'' cycle. 
As P\textsubscript{4}-EE noted ``watching the video took too much time''.

The lower documentation clarity scores for v4.1 in Study 1 paradoxically highlight a mechanical success. 
Despite an alpha-stage guide that was deliberately thinned (§\ref{sec:deliver}), the 30-point mean SUS gain (Table \ref{tab:metrics}; Fig.~\ref{fig:radars-Sus_docs}a) suggests that the hardware affordances carried the load previously managed by dense text. 
As P\textsubscript{3}-EE  observed ``I [feel] I can assemble everything without text. 
And if I am stuck or confused, then I go to read [the docs].''

\subsection{Limitations}
Both studies used small, non-random samples: Study 1 from the development team and Study 2 from partner institutions in Guyana and Estonia.
Study~1 in particular included two senior-participants among its four developers, introducing self-evaluation bias that we have flagged where it most directly affects interpretation.
The Study~1 task order may have inflated v4 performance through practice transfer (§\ref{sec:study1}); Study~2 tested only v4.1, so user-side gains cannot be quantified as a within-subjects delta. 
Because we deliberately thinned the v4.1 documentation, cross-version doc clarity scores in Study~1 do not fully reflect documentation quality.
Lastly, thematic coding lacked inter-rater reliability checks, and single-session study designs cannot capture long-term attachment or perceived agency. 

\section{Conclusion}
We contribute an end-to-end co-design methodology, and empirical evidence that DfA and DfD principles can transform open-source social robot assembly from a barrier into an accessible entry point. 
The key insight: \textit{perceived} difficulty governs whether students attempt and complete a build, and DfA/DfD redesign reshapes how the build feels before it shortens how long it takes. 
Remaining friction concentrates in wiring identification and documentation format, addressable through component labelling and hybrid documentation that combines video, still images and step-by-step diagrams. 
More broadly, treating buildability and serviceability as constituent elements of the human--robot relationship reframes assembly as the opening move of social interaction, opening an agenda linking material co-creation to attachment, ownership, and sustained engagement. 
We release all open-source artefacts at \url{https://rsc.ee/} and welcome contributions. 

\begin{credits}
\subsubsection{\ackname}
We sincerely thank all the students in Guyana and Estonia for their enthusiastic participation and feedback, which were essential to this project. 
This work was in part co-funded by the European Union and the Estonian Research Council grant (PRG3237).

\subsubsection{Author Contributions (CRediT).}
\textbf{F. Baksh:} Conceptualisation, Methodology, Investigation (Estonia case study, Study~1 facilitation), Validation,  Data curation, Resources, Project administration, Writing -- original draft -- review \& editing. 
\textbf{M.B. Zorec:} Conceptualisation, Methodology, Software, Validation, Formal analysis, Resources, Data curation, Visualisation, Writing -- review \& editing. 
\textbf{F. Baksh:} Investigation (Guyana case study, Study~2 facilitation), Writing -- review \& editing. 
\textbf{K. Kruusam\"ae:} Supervision, Funding acquisition, Writing -- review \& editing.

\subsubsection{Use of Generative AI Tools.}
The authors used several generative AI assistants for language editing and copy-editing. 
The authors performed all data analysis, interpretation and citation verification, and take full responsibility for the manuscript's content.
\end{credits}
\bibliographystyle{splncs04}
\bibliography{references}
\end{document}